\documentclass{article}

\usepackage{microtype}
\usepackage{graphicx}
\usepackage{subcaption}
\usepackage{booktabs} 

\usepackage{hyperref}

\usepackage{booktabs}
\usepackage{amsmath}
\usepackage{xcolor}
\usepackage{amssymb}
\usepackage{subcaption}
\usepackage{graphicx}
\usepackage{algorithm}
\usepackage{algorithmic}

\usepackage[accepted]{eiml_icml2026}

\usepackage{amsmath}
\usepackage{amssymb}
\usepackage{mathtools}
\usepackage{amsthm}

\usepackage[capitalize,noabbrev]{cleveref}

\theoremstyle{plain}
\newtheorem{theorem}{Theorem}[section]
\newtheorem{proposition}[theorem]{Proposition}

\theoremstyle{definition}

\theoremstyle{remark}

\usepackage[textsize=tiny]{todonotes}

\icmltitlerunning{Stacking-Based Model Weighting in Large Language Bayes}

\begin{document}

\twocolumn[
  \icmltitle{Stacking-Based Weighting for Large Language Bayes in M-Open Settings}

  \icmlsetsymbol{equal}{*}

  \begin{icmlauthorlist}
    \icmlauthor{Anant Bhide}{umass}
    \icmlauthor{Akarsh Gupta}{umass}
    \icmlauthor{Edmond Cunningham}{umass}
    \icmlauthor{Justin Domke}{umass}
  \end{icmlauthorlist}

  \icmlaffiliation{umass}{Manning College of Information and Computer Sciences, University of Massachusetts Amherst, Amherst, MA, USA}

  \icmlcorrespondingauthor{Anant Bhide}{abhide@umass.edu}
  \icmlcorrespondingauthor{Akarsh Gupta}{akarshgupta@umass.edu}
  \icmlcorrespondingauthor{Edmond Cunningham}{edmondcunnin@cs.umass.edu}
  \icmlcorrespondingauthor{Justin Domke}{domke@cs.umass.edu}

  \icmlkeywords{Machine Learning, ICML}

  \vskip 0.3in
]

\printAffiliationsAndNotice{}  
\begin{abstract}

Large Language Bayes (LLB) transforms a natural language problem description into a collection of candidate probabilistic programs and performs inference within each model, combining their predictions using Bayesian model averaging. While this provides a flexible framework for probabilistic inference without manual model specification, the resulting candidate model set is often misspecified in practice, and likelihood-based aggregation can concentrate on a single model, under-representing uncertainty across the sampled models. We propose a stacking-based aggregation method for LLB that selects model weights by optimizing predictive performance rather than marginal likelihood, while preserving the original language-model-driven model generation and within-model inference pipeline. Because leave-one-out predictive quantities are intractable in latent-variable models, we derive a variational lower bound that yields a tractable ELBO-based stacking objective. Across real-world probabilistic inference tasks, our method achieves comparable predictive performance to Bayesian model averaging while yielding more dispersed model weights, reducing concentration on a single candidate model under misspecification.
\end{abstract}

\section{Introduction}

Probabilistic inference is a fundamental tool for reasoning under uncertainty and making decisions from data. In practice, however, its performance depends critically on the chosen model, and specifying an appropriate probabilistic model is often the hardest part of the inference pipeline. This difficulty is especially pronounced in realistic settings, where the true data-generating process is unknown and any candidate model class is necessarily incomplete.

Recent work on Large Language Bayes (LLB) \cite{domke2025large} aims to reduce the burden of model specification by using a language model to generate candidate probabilistic programs from a textual description of the problem. Given a textual description of the task and observed data, LLB samples a collection of candidate models, performs inference within each model, and combines the resulting posterior distributions over target variables using Bayesian model averaging. This yields a flexible framework for probabilistic inference without requiring the user to manually specify a single exact model.

However, LLB is subject to model misspecification, where the set of candidate models may not include the true data-generating process, a setting referred to as $M$-open \cite{yao2017stacking}. In this regime, Bayesian model averaging can behave more like model selection than model combination, concentrating its mass on a small subset of candidate models. In the context of LLB, this can lead to a collapse of the ensemble onto a single misspecified model, even when multiple models provide comparable predictive performance.

In this paper, we replace evidence-based weighting with stacking \cite{yao2017stacking}. Rather than interpreting model weights as posterior probabilities, stacking selects weights to maximize out-of-sample predictive performance. This makes stacking well suited to the $M$-open regime induced by language-generated model classes, where the goal is not to recover a true model but to construct a predictive ensemble that remains robust under misspecification.

Because LLB operates with latent-variable models, the leave-one-out predictive quantities required by stacking are generally intractable. We therefore derive a tractable approximation based on a variational lower bound, allowing the stacking objective to be implemented using the same inference machinery already used within LLB. The resulting method preserves the model-generation and within-model inference pipeline of LLB while replacing its aggregation rule with a predictive weighting scheme that avoids degenerate concentration on a single model.

In summary, we propose a stacking-based aggregation method for LLB that replaces Bayesian model averaging with a leave-one-out predictive objective, together with a variational approximation that makes this objective tractable in latent-variable settings, enabling more reliable aggregation under model uncertainty in language-generated model spaces.

\section{Background}

This work builds on two main components: Large Language Bayes and stacking-based predictive model averaging. We introduce only the minimal setup required to describe our method.

\subsection{Large Language Bayes}
Large Language Bayes (LLB) \cite{domke2025large} is an automated approach for performing probabilistic inference over latent variables given data and a textual description of the problem.
Given a prompt \(t\) describing the problem and observed data \(x = (x_1, \dots, x_n)\), LLB samples a collection of candidate models
\[
m^{(k)} \sim p(m \mid t), \quad k = 1, \dots, K.
\]

Each model \(m^{(k)}\) defines a joint distribution over latent variables \(z\) and observations \(x\). Inference is performed independently within each model to obtain posterior distributions \(p(z \mid x, m^{(k)}, t)\).

The overall objective is to approximate the posterior over latent variables by marginalizing over models:
\[
p(z \mid x, t) = \sum_{k=1}^K p(m^{(k)} \mid x, t)\ p(z \mid x, m^{(k)})\
\]

In practice, sampling from \(p(m \mid x, t)\) is intractable. Instead, LLB samples models from \(p(m \mid t)\) and assigns importance weights based on the model evidence. This yields the approximation
\[
p(z \mid x, t) \approx \sum_{k=1}^K w_k^{\mathrm{BMA}}\, p(z \mid x, m^{(k)})
\]
where $w_k^{\mathrm{BMA}} \propto p(x \mid m^{(k)})$ are the standard Bayesian model averaging weights proportional to the marginal likelihood of each candidate model.

This construction corresponds to Bayesian model averaging over the sampled candidate models and therefore inherits the usual closed-world interpretation that one of the candidate models is expected to be adequate.

\subsection{The \(M\)-open setting}
In many practical scenarios, including LLB, the true data-generating process is not contained within the set of candidate models. This is known as the \(M\)-open setting, where the objective is not to recover a true model, but to obtain accurate predictions \cite{le2017bayes, Clydec2013}.

In this regime, Bayesian model averaging can concentrate its weight on a small subset of candidate models, effectively behaving more like model selection than model combination \cite{yao2017stacking,le2017bayes}.

Since LLB generates models using a language model, there is no reason to expect that the true model lies within the candidate set. Therefore, LLB naturally operates in an \(M\)-open regime, where likelihood-based weighting can lead to undesirable weight concentration on a single misspecified model.

\subsection{Stacking and predictive model averaging}
Stacking provides an alternative framework for combining models that is explicitly designed for predictive performance. Rather than interpreting weights as posterior probabilities of models, stacking selects weights to maximize out-of-sample predictive accuracy under a leave-one-out scheme, typically measured using a proper scoring rule such as the log score \cite{yao2017stacking,vehtari2017practical}.

This leads to weights that directly target predictive accuracy rather than model identification. This predictive perspective makes stacking a natural replacement for the weighting scheme used in LLB, aligning the aggregation of models with the goal of accurate prediction under model misspecification.

\section{Related Work}

Our work sits at the intersection of language-conditioned probabilistic modeling and predictive aggregation under model misspecification.

Recent work has explored the use of natural language as a specification medium for probabilistic inference and prediction. \cite{wong2023wordmodels} translates text into probabilistic programs and performs Bayesian inference in the resulting model, while \cite{requeima2024llmprocesses} uses language to condition numerical predictive distributions directly. \cite{saad2019bayesian} and \cite{li2024automated} further develop the idea of generating probabilistic programs or statistical models automatically from data and natural language. Large Language Bayes \cite{domke2025large} builds on this line by treating a text description as inducing a distribution over formal probabilistic models and then performing inference within each sampled model. In this paper, we take the model-generation step of LLB as given and focus instead on how the resulting collection of candidate models should be aggregated.

A separate line of work studies how to combine predictive models when the true data-generating process is not assumed to be in the candidate class. Bayesian model averaging is the classical framework for accounting for model uncertainty \cite{hoeting1999bma,wasserman2000bmsma}, but in the $M$-open setting it can concentrate too strongly on a small subset of candidate models and fail to exploit complementary predictive structure. Stacking addresses this issue by selecting weights that maximize out-of-sample predictive performance rather than posterior model probability \cite{yao2017stacking,vehtari2017practical}. Le and Clarke \cite{le2017bayes} provide a Bayesian decision-theoretic interpretation of stacking in both $M$-complete and $M$-open settings, showing that stacking weights can be viewed as asymptotically minimizing posterior expected loss. Related extensions such as Bayesian hierarchical stacking further generalize stacking to input-dependent, partially pooled model weights \cite{yao2021hierarchical}. In practice, the required leave-one-out predictive quantities can be efficiently approximated using Pareto-smoothed importance sampling \cite{vehtari2017practical}.

Our setting combines these two perspectives. LLB provides a way to generate a finite set of candidate probabilistic programs from language, but the induced model class should not be assumed to contain the true data-generating process. We therefore replace evidence-based aggregation with a stacking objective defined over leave-one-out predictive approximations. This yields a predictive combination rule that is aligned with the $M$-open nature of language-generated model classes, while preserving the latent-variable structure and within-model inference machinery of LLB.

\section{Method}
\label{sec:method}

We retain the LLB pipeline for model generation and inference \cite{domke2025large}, but replace the original evidence-based weighting rule with a regularized stacking-based objective that selects model weights according to predictive performance \cite{yao2017stacking,vehtari2017practical}.

Figure~\ref{fig:llb_pipeline_with_stacking} illustrates the LLB pipeline together with the proposed stacking-based aggregation.

\subsection{Stacking-Based Weighting Objective}

Given a prompt $t$, LLB samples a finite set of candidate models
\[
m^{(k)} \sim p(m \mid t), \quad k = 1, \dots, K.
\]

For each candidate model $m^{(k)}$, and for each observation $x_i$, we consider the leave-one-out predictive density
\[
p(x_i \mid x_{-i}, m^{(k)}, t),
\]
where $x_{-i}$ denotes the dataset with the $i$-th observation removed.

We define weights $w = (w_1, \dots, w_K)$ constrained to the probability simplex,
\[
w_k \ge 0, \quad \sum_{k=1}^K w_k = 1.
\]

The corresponding stacked predictive distribution for the held-out observation $x_i$ is
\[
\sum_{k=1}^K w_k \, p(x_i \mid x_{-i}, m^{(k)}, t).
\]

Following the stacking framework, we choose the weights by maximizing the aggregate leave-one-out log predictive score. To preserve the inductive bias encoded by the original LLB weighting rule, we add a KL regularizer that anchors the learned weights near the evidence-based BMA weights while still allowing the stacking objective to redistribute mass when this improves predictive performance:
\[
\begin{aligned}
\hat{w}
=
\arg\max_{w \in \Delta^K}
\Bigg[
& \sum_{i=1}^n
\log \left(
\sum_{k=1}^K w_k \, p(x_i \mid x_{-i}, m^{(k)}, t)
\right) \\
& - \lambda \, \mathrm{KL}\!\left(w^{\mathrm{BMA}} \,\|\, w\right)
\Bigg].
\end{aligned}
\]

This objective directly targets predictive performance under an $M$-open setting, where the true data-generating process is not assumed to lie in the candidate model class.

However, the quantities $p(x_i \mid x_{-i}, m^{(k)}, t)$ are generally intractable in the LLB setting because of latent variables. We therefore approximate these terms using a variational bound. The key result is summarized in the following proposition, with proof deferred to Appendix~\ref{app:proof_elbo_stacking}.

\begin{proposition}[ELBO-based Stacking Objective]
\label{prop:elbo_stacking}
For each candidate model $m^{(k)}$ and observation $x_i$, define a variational distribution
$q_{i,k}(z)$ approximating the posterior
$p(z \mid x_i, x_{-i}, m^{(k)}, t)$, and define
\[
\mathrm{ELBO}_{i,k}
=
\mathbb{E}_{q_{i,k}(z)}
\Big[
\log p(x_i, z \mid x_{-i}, m^{(k)}, t)
\;-\;
\log q_{i,k}(z)
\Big].
\]
Then the leave-one-out predictive density satisfies
\[
\mathrm{ELBO}_{i,k} \le \log p(x_i \mid x_{-i}, m^{(k)}, t),
\]
and substituting this variational bound into the stacking objective yields
\[
\begin{aligned}
\hat{w}
=
\arg\max_{w \in \Delta^K}
\Bigg[
& \sum_{i=1}^n
\log \left(
\sum_{k=1}^K w_k \exp(\mathrm{ELBO}_{i,k})
\right) \\
& - \lambda \, \mathrm{KL}\!\left(w^{\mathrm{BMA}} \,\|\, w\right)
\Bigg].
\end{aligned}
\]
\end{proposition}

This is the objective we optimize in practice. It preserves the structure of stacking while replacing each intractable leave-one-out predictive term with a variational approximation that can be computed using the same inference machinery already used within LLB.

\subsection{Full Algorithm}

We now summarize the complete inference and weighting pipeline combining LLB-style model generation, variational leave-one-out estimation, stacking-based weight optimization, and posterior aggregation.

\begin{algorithm}[h]
\caption{LLB with KL-Regularized Stacking-Based ELBO Weighting}
\label{alg:llb-stacking}
\begin{algorithmic}[1]

\REQUIRE Text description $t$, dataset $x = (x_1, \dots, x_n)$, number of models $K$

\STATE Sample models:
$ m^{(k)} \sim p(m \mid t), \quad k = 1, \dots, K $

\FOR{each model $k = 1, \dots, K$}
    \FOR{each data point $i = 1, \dots, n$}
        \STATE Form leave-one-out dataset $x_{-i}$
        \STATE Construct variational posterior:
        $ q_{i,k}(z) \approx p(z \mid x_i, x_{-i}, m^{(k)}, t) $
        \STATE Compute ELBO-based surrogate predictive score:
        $
        \mathrm{ELBO}_{i,k}
        =
        \mathbb{E}_{q_{i,k}(z)}
        \left[
        \log \frac{p(x_i, z \mid x_{-i}, m^{(k)}, t)}{q_{i,k}(z)}
        \right]
        $
    \ENDFOR
\ENDFOR

\STATE Solve stacking optimization:
\[
\begin{aligned}
\hat{w}
=
\arg\max_{w \in \Delta^K}
\Bigg[
& \sum_{i=1}^n
\log \left(
\sum_{k=1}^K w_k \exp(\mathrm{ELBO}_{i,k})
\right) \\
& - \lambda \, \mathrm{KL}\!\left(w^{\mathrm{BMA}} \,\|\, w\right)
\Bigg].
\end{aligned}
\]

\STATE Compute final posterior mixture:
$
\tilde{p}(z \mid x, t)
=
\sum_{k=1}^K \hat{w}_k \, p(z \mid x, m^{(k)}, t)
$

\RETURN $ \tilde{p}(z \mid x, t), \hat{w} $

\end{algorithmic}
\end{algorithm}

\begin{figure*}[t]
    \centering
    \includegraphics[width=0.9\textwidth]{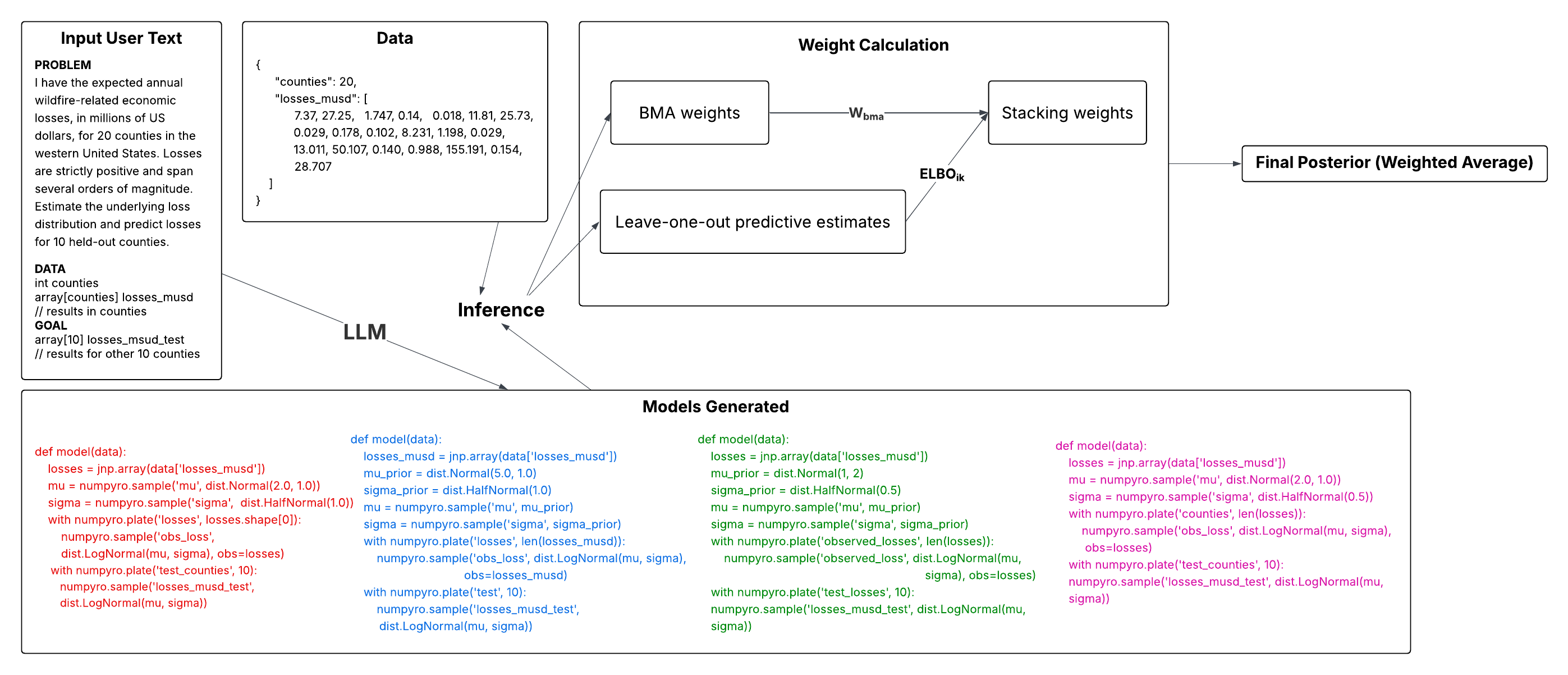}
    \caption{Example LLB pipeline with a language prompt, sampled candidate models, and ELBO-based stacking with KL regularization toward BMA weights.}
    \label{fig:llb_pipeline_with_stacking}
\end{figure*}

\section{Experiments}

\subsection{Experimental Setup}

We evaluate the proposed KL-regularized ELBO-stacking weighting scheme against the original LLB aggregation rule on a suite of probabilistic inference tasks. In line with the LLB framework, each task is specified by a textual problem description $t$ together with observed data $x$, while held-out observations are reserved for predictive evaluation.

We consider three county-level expected annual loss prediction tasks covering different natural hazard regimes: wildfire, inland flooding, and hurricane, derived from the FEMA National Risk Index dataset \cite{fema_nri}. For each task, we construct a training set of $n=20$ observed county-level loss values (in millions of USD), and evaluate predictive performance on 10 held-out counties. Across all tasks, losses are strictly positive and exhibit heavy-tailed variability.

For each task, candidate probabilistic programs are generated from the textual description $t$ following the LLB pipeline. Figure~\ref{fig:llb_pipeline_with_stacking} illustrates the full pipeline from language prompt to posterior aggregation.

\subsection{Method Details}

For each task, we generate 250 candidate NumPyro programs using Qwen2.5 Coder 32B. Programs are generated from the textual description $t$ alone, deduplicated by normalized code hash, and filtered to retain only those that compile and pass shape validation.

Each valid candidate model is fit using the No-U-Turn Sampler (NUTS) \cite{hoffman2014nuts}, a gradient-based Markov Chain Monte Carlo method, producing posterior samples for downstream evaluation.

To approximate the leave-one-out predictive densities required for stacking, we construct $x_{-i}$ for each observation and compute a variational lower bound (ELBO) on $\log p(x_i \mid x_{-i}, m^{(k)}, t)$ using a Gaussian variational family. These ELBO values provide a tractable surrogate for the leave-one-out predictive objective and form the basis of weight optimization.

We compare the original LLB evidence-based aggregation rule with the KL-regularized stacking objective introduced in Section~\ref{sec:method}. The KL term stabilizes optimization by encouraging proximity to the BMA weights while allowing deviations when supported by predictive performance. We select $\lambda$ from a small grid using validation predictive performance on a held-out validation split. We optimize this objective in log-space using SciPy's BFGS with softmax parameterization ($w = \text{softmax}(\theta)$) and analytical gradients.

\subsection{Evaluation Metrics}

We evaluate both the distribution of learned model weights and predictive performance. To assess weight concentration, we report the effective sample size (ESS), defined as $1 / \sum_k w_k^2$, where values close to 1 indicate collapse to a single model and larger values indicate more distributed weighting. We also report the top model weights to directly visualize how mass is allocated across models.

Predictive performance is measured using expected log predictive density (ELPD) on held-out test data \cite{vehtari2017practical}, where higher values indicate better generalization. This allows us to verify that distributing weight across models does not degrade predictive performance.

\subsection{Results}

\begin{figure}[t]
    \centering
    \includegraphics[width=\linewidth]{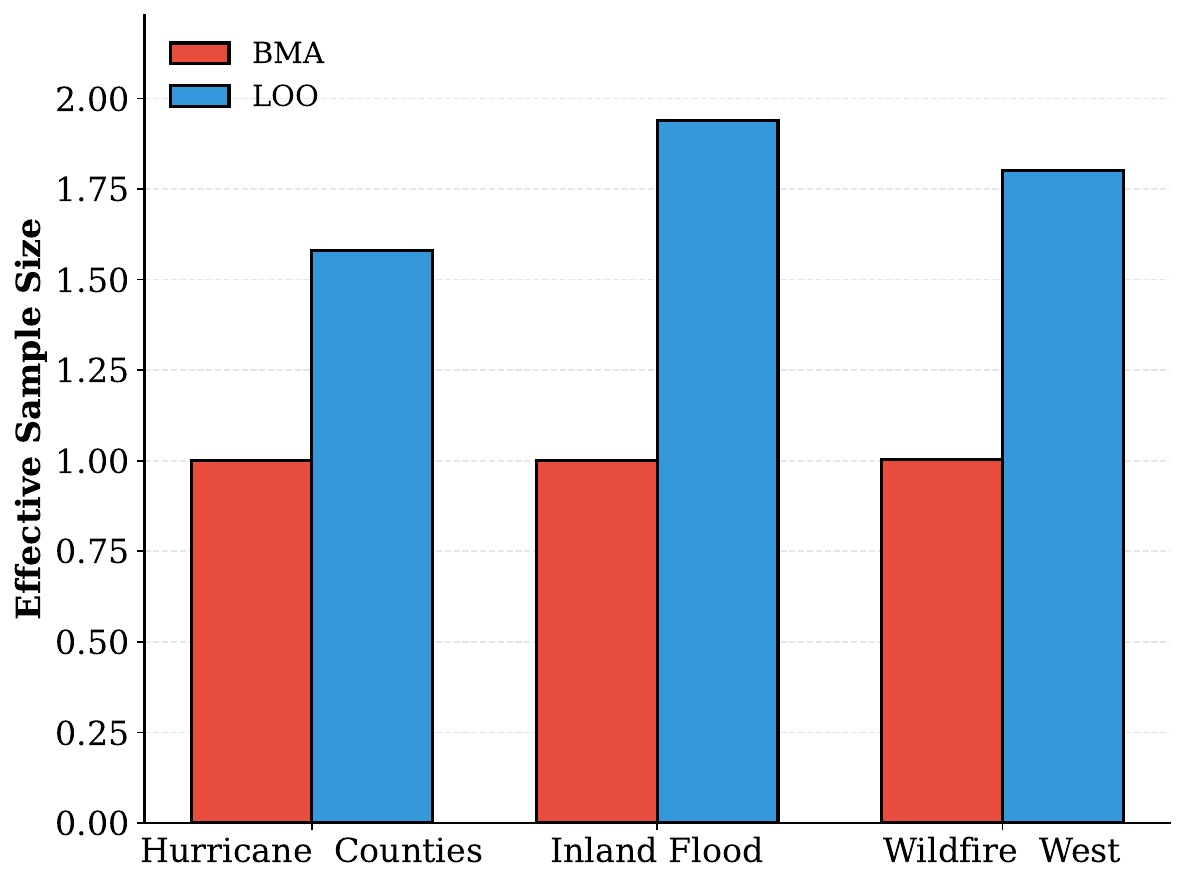}
    \caption{Effective sample size (ESS) of learned model weights across tasks.}
    \label{fig:ess}
\end{figure}

Figure~\ref{fig:ess} and Table~\ref{tab:weights_all} show that BMA concentrates most of its mass on a single model, whereas stacking spreads weight across multiple candidate models.

This behavior is consistent across all three hazard regimes. In the BMA case, the posterior over models effectively collapses to a single candidate, indicating that evidence-based weighting behaves like model selection in practice. In contrast, stacking assigns non-trivial weight to multiple models, reflecting residual uncertainty over which misspecified model best approximates the data-generating process in the $M$-open setting.

\begin{table}[h]
\caption{Top model weights under BMA and stacking.}
\label{tab:weights_all}
\centering
\begin{tabular}{l l l}
\toprule
Dataset & BMA top weights & Stacking top weights \\
\midrule

\textbf{Hurricane} 
& \begin{tabular}[t]{@{}l@{}}
$w_{139}=1.000$
\end{tabular}
&
\begin{tabular}[t]{@{}l@{}}
$w_{1}=0.759$ \\
$w_{65}=0.238$ \\
$w_{129}=0.003$
\end{tabular}
\\

\midrule

\textbf{Inland Flood} 
& \begin{tabular}[t]{@{}l@{}}
$w_{24}=1.000$
\end{tabular}
&
\begin{tabular}[t]{@{}l@{}}
$w_{24}=0.588$ \\
$w_{13}=0.412$
\end{tabular}
\\

\midrule

\textbf{Wildfire} 
& \begin{tabular}[t]{@{}l@{}}
$w_{83}=0.998$
\end{tabular}
&
\begin{tabular}[t]{@{}l@{}}
$w_{138}=0.666$ \\
$w_{83}=0.334$
\end{tabular}
\\

\bottomrule
\end{tabular}
\end{table}

Table~\ref{tab:predictive} reports held-out predictive performance. Across all tasks, stacking matches Bayesian model averaging in terms of ELPD, indicating that redistributing weight across multiple models does not degrade predictive accuracy. This supports the interpretation that the primary difference between the two approaches is not predictive power, but rather the representation of model uncertainty.

\begin{table}[h]
\caption{Held-out predictive performance measured by expected log predictive density (ELPD).}
\label{tab:predictive}
\centering
\begin{tabular}{lrr}
\toprule
Dataset & LLB (BMA) & Stacking \\
\midrule
Hurricane & -5.983 & -5.999 \\
Inland Flood & -370.340 & -370.340 \\
Wildfire & -177.922 & -177.897 \\
\bottomrule
\end{tabular}
\end{table}

\section{Discussion and Limitations}

Our results suggest that replacing evidence-based aggregation in LLB with a stacking-based objective yields weight distributions that are more consistent with the $M$-open setting. In particular, stacking produces less concentrated weights than Bayesian model averaging while maintaining comparable predictive performance. This supports our main claim: in language-generated model spaces, the goal is not to identify a single dominant model, but to maintain a meaningful mixture over plausible misspecified candidates. At the same time, our experiments do not establish that the learned weights recover the true epistemic uncertainty of the data-generating process. The resulting mixture still depends on the quality and diversity of candidate models generated by the language model. Future work includes evaluating the approach on more diverse problems and exploring alternative regularization strategies and input-dependent stacking schemes.

\section{Conclusion}

We proposed a stacking-based aggregation method for Large Language Bayes that replaces evidence-based weighting with a predictive objective while preserving the original language-model-driven model generation and within-model inference pipeline. Across our experiments, the method maintains comparable predictive performance to the LLB baseline while yielding a more distributed allocation of weights across candidate models. These results support the view that stacking is a natural fit for LLB-style inference in the $M$-open setting, where the goal is robust prediction and a more faithful representation of model uncertainty under misspecification.

\bibliography{bibliography}
\bibliographystyle{icml2026}

\newpage




\onecolumn

\appendix
\section{Appendix}

\subsection{Proof of Proposition~\ref{prop:elbo_stacking}}
\label{app:proof_elbo_stacking}

\begin{proof}

Fix a candidate model $m^{(k)}$, an observation $x_i$, the remaining data $x_{-i}$, and the prompt $t$. Let $z$ denote the latent variables in the model, and let $q_{i,k}(z)$ be any variational distribution over $z$.

We begin with the leave-one-out predictive density:
\begin{align}
p(x_i \mid x_{-i}, m^{(k)}, t)
&=
\int p(x_i, z \mid x_{-i}, m^{(k)}, t)\, dz.
\label{eq:app_loo_marginal}
\end{align}

We consider the Kullback--Leibler divergence from $q_{i,k}(z)$ to the leave-one-out posterior:
\begin{align}
\mathrm{KL}\!\left(q_{i,k}(z) \,\|\, p(z \mid x_i, x_{-i}, m^{(k)}, t)\right)
&=
\int q_{i,k}(z)
\log \frac{q_{i,k}(z)}{p(z \mid x_i, x_{-i}, m^{(k)}, t)}
\, dz.
\label{eq:app_kl_def}
\end{align}

By Bayes' rule,
\begin{align}
p(z \mid x_i, x_{-i}, m^{(k)}, t)
&=
\frac{p(x_i, z \mid x_{-i}, m^{(k)}, t)}
     {p(x_i \mid x_{-i}, m^{(k)}, t)}.
\label{eq:app_bayes_rule}
\end{align}

Substituting \eqref{eq:app_bayes_rule} into \eqref{eq:app_kl_def} gives
\begin{align}
\mathrm{KL}\!\left(q_{i,k}(z) \,\|\, p(z \mid x_i, x_{-i}, m^{(k)}, t)\right)
&=
\int q_{i,k}(z)
\log \frac{q_{i,k}(z)\, p(x_i \mid x_{-i}, m^{(k)}, t)}
{p(x_i, z \mid x_{-i}, m^{(k)}, t)}
\, dz.
\label{eq:app_kl_sub}
\end{align}

Since $p(x_i \mid x_{-i}, m^{(k)}, t)$ does not depend on $z$, we may separate it from the integral:
\begin{align}
\mathrm{KL}\!\left(q_{i,k}(z) \,\|\, p(z \mid x_i, x_{-i}, m^{(k)}, t)\right)
&=
\int q_{i,k}(z)
\log \frac{q_{i,k}(z)}{p(x_i, z \mid x_{-i}, m^{(k)}, t)}
\, dz
\nonumber \\
&\quad +
\log p(x_i \mid x_{-i}, m^{(k)}, t).
\label{eq:app_kl_expand}
\end{align}

Rearranging terms yields
\begin{align}
\log p(x_i \mid x_{-i}, m^{(k)}, t)
&=
\int q_{i,k}(z)
\log \frac{p(x_i, z \mid x_{-i}, m^{(k)}, t)}{q_{i,k}(z)}
\, dz
\nonumber \\
&\quad +
\mathrm{KL}\!\left(q_{i,k}(z) \,\|\, p(z \mid x_i, x_{-i}, m^{(k)}, t)\right).
\label{eq:app_elbo_decomp}
\end{align}

We define the evidence lower bound:
\begin{align}
\mathrm{ELBO}_{i,k}
:=
\int q_{i,k}(z)
\log \frac{p(x_i, z \mid x_{-i}, m^{(k)}, t)}{q_{i,k}(z)}
\, dz
=
\mathbb{E}_{q_{i,k}(z)}
\left[
\log \frac{p(x_i, z \mid x_{-i}, m^{(k)}, t)}{q_{i,k}(z)}
\right].
\label{eq:app_elbo_def}
\end{align}

Therefore,
\begin{align}
\log p(x_i \mid x_{-i}, m^{(k)}, t)
=
\mathrm{ELBO}_{i,k}
+
\mathrm{KL}\!\left(q_{i,k}(z) \,\|\, p(z \mid x_i, x_{-i}, m^{(k)}, t)\right).
\label{eq:app_final_decomp}
\end{align}

Because the Kullback--Leibler divergence is non-negative,
\begin{align}
\mathrm{KL}\!\left(q_{i,k}(z) \,\|\, p(z \mid x_i, x_{-i}, m^{(k)}, t)\right) \ge 0,
\end{align}
it follows that
\begin{align}
\mathrm{ELBO}_{i,k} \le \log p(x_i \mid x_{-i}, m^{(k)}, t).
\label{eq:app_elbo_bound}
\end{align}

Applying the exponential function, which is monotone increasing, preserves the inequality:
\begin{align}
\exp(\mathrm{ELBO}_{i,k})
\le
p(x_i \mid x_{-i}, m^{(k)}, t).
\label{eq:app_exp_bound}
\end{align}

Define the surrogate predictive density
\begin{align}
\tilde{p}_{i,k} := \exp(\mathrm{ELBO}_{i,k}).
\label{eq:app_surrogate_def}
\end{align}
Then \eqref{eq:app_exp_bound} implies
\begin{align}
\tilde{p}_{i,k} \le p(x_i \mid x_{-i}, m^{(k)}, t).
\end{align}

Substituting this variational bound into the stacking objective gives the surrogate objective
\begin{align}
\hat{w}
&=
\arg\max_{w \in \Delta^K}
\Bigg[
\sum_{i=1}^n
\log \left(
\sum_{k=1}^K w_k \tilde{p}_{i,k}
\right)
- \lambda \, \mathrm{KL}\!\left(w^{\mathrm{BMA}} \,\|\, w\right)
\Bigg]
\nonumber \\
&=
\arg\max_{w \in \Delta^K}
\Bigg[
\sum_{i=1}^n
\log \left(
\sum_{k=1}^K w_k \exp(\mathrm{ELBO}_{i,k})
\right)
- \lambda \, \mathrm{KL}\!\left(w^{\mathrm{BMA}} \,\|\, w\right)
\Bigg].
\label{eq:app_final_objective}
\end{align}

This proves Proposition~\ref{prop:elbo_stacking}.

\end{proof}

\subsection{ELBO Convergence Diagnostics}
\label{app:elbo_convergence}

\begin{figure}[h]
    \centering
    \includegraphics[width=\textwidth]{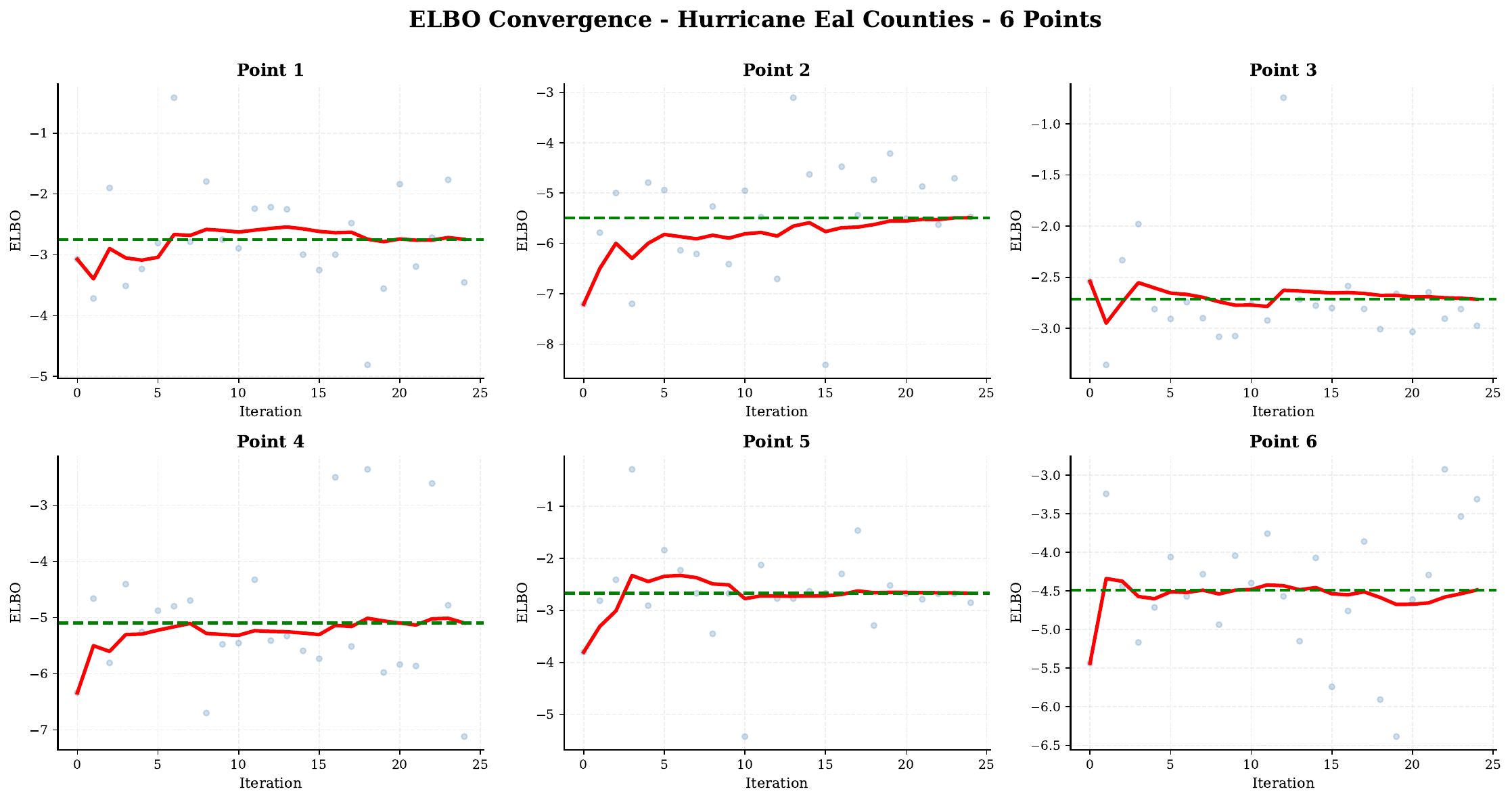}
    \caption{ELBO convergence for 6 sample points from the Hurricane task. Each subplot shows the optimization trajectory for a single leave-one-out inference problem.}
    \label{fig:elbo_hurricane}
\end{figure}

\begin{figure}[h]
    \centering
    \includegraphics[width=\textwidth]{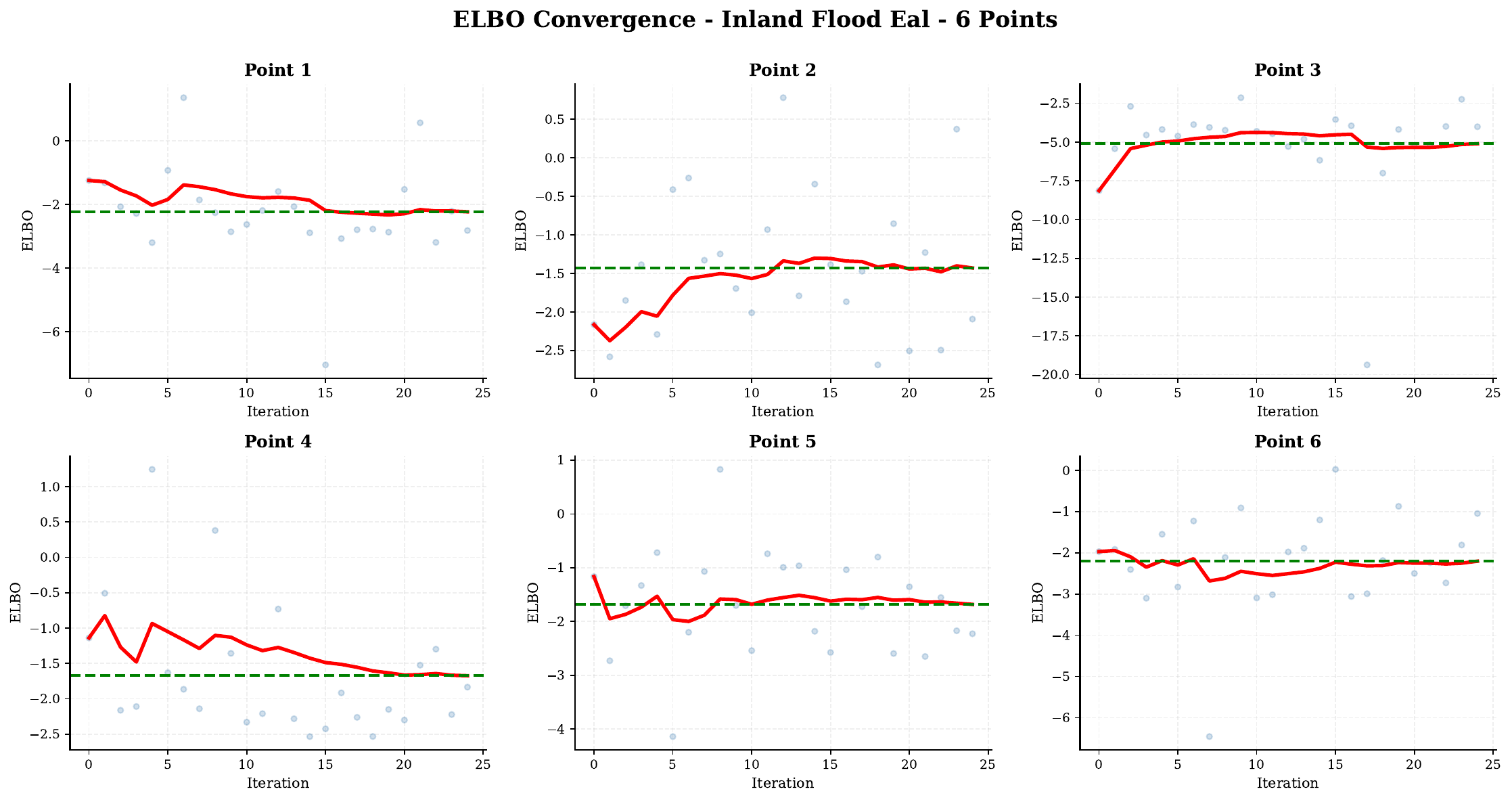}
    \caption{ELBO convergence for 6 sample points from the Inland Flood task.}
    \label{fig:elbo_flood}
\end{figure}

\begin{figure}[h]
    \centering
    \includegraphics[width=\textwidth]{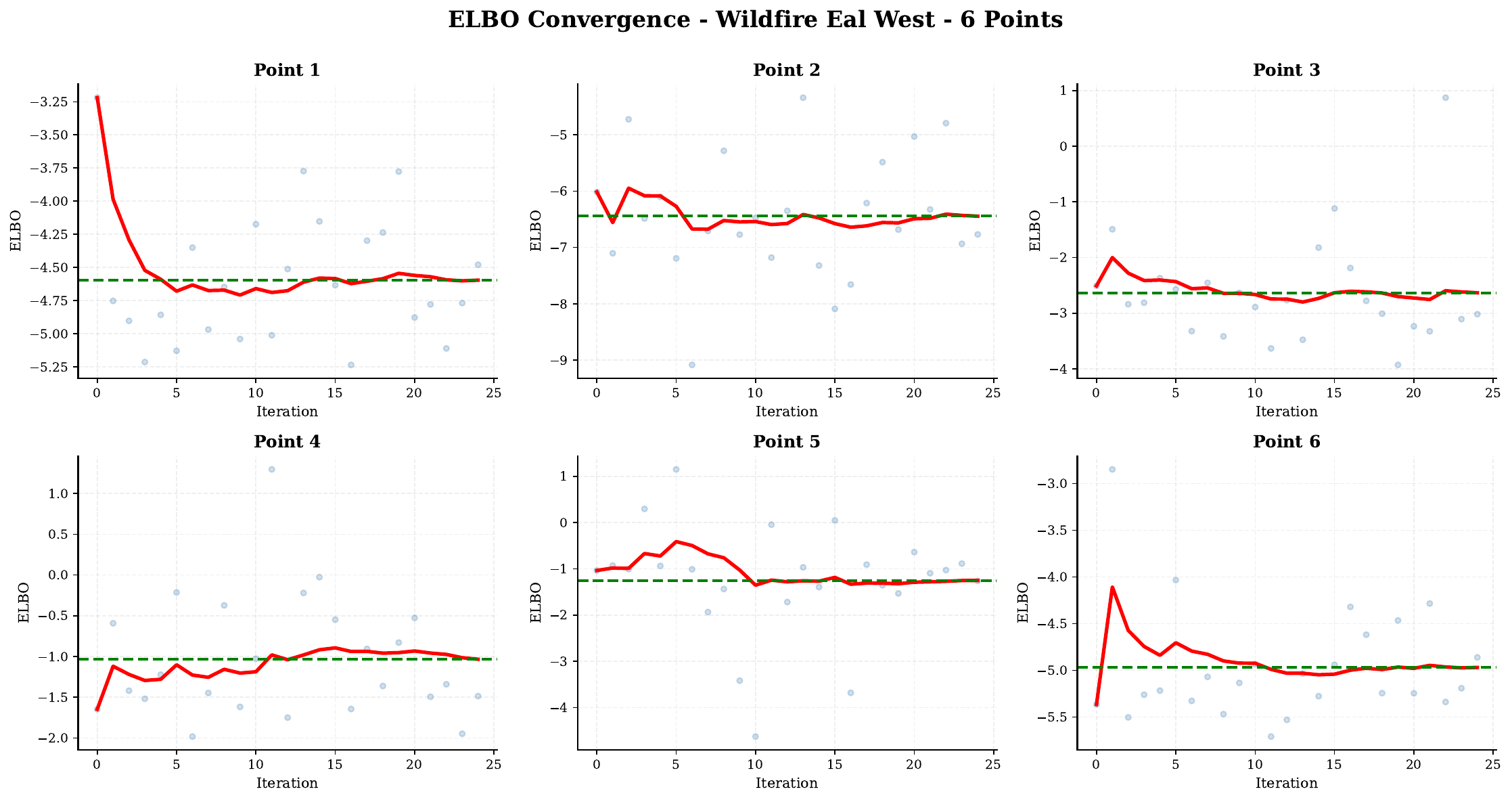}
    \caption{ELBO convergence for 6 sample points from the Wildfire task.}
    \label{fig:elbo_wildfire}
\end{figure}

To verify the quality of the variational approximations used in our stacking objective, we track the convergence of the ELBO values during optimization for each leave-one-out inference problem. Figures~\ref{fig:elbo_hurricane}, \ref{fig:elbo_flood}, and \ref{fig:elbo_wildfire} show ELBO trajectories for a sample of 6 data points from each of the three tasks.

We construct a variational approximation $q_{i,k}(z)$ and optimize the corresponding ELBO objective using gradient-based variational inference. The red solid lines show the ELBO values at each iteration, while the green dashed lines indicate the final converged values used in the stacking objective.

Across all three tasks, the ELBO values converge within 25 iterations, typically stabilizing after 10--15 iterations. The gray points in the background represent the raw ELBO evaluations during optimization. The convergence patterns suggest that the variational approximation provides stable and reliable surrogates for the intractable leave-one-out predictive densities required by the stacking objective.

\subsection{Code}
The link to the code is: \href{https://github.com/Akarsh-dundun/Large-Language-Bayes-Model}{Code}

\end{document}